\documentclass[letterpaper, 10 pt, conference]{ieeeconf}
\IEEEoverridecommandlockouts
\usepackage[english]{babel}
\usepackage{hyperref}

\usepackage{xcolor}
\usepackage{colortbl}
\usepackage[utf8]{inputenc}
\usepackage{graphicx}
\usepackage{multirow}
\usepackage{tabularx}
\usepackage{balance}
\usepackage{cite}

\usepackage{subcaption}
\usepackage{multicol}
\usepackage{cuted}
\usepackage{booktabs}
\usepackage{array}
\usepackage{float}
\usepackage{caption}
\usepackage{amsmath}
\usepackage{amsfonts}
\usepackage{amssymb}
\usepackage{arydshln}
\usepackage[english]{cleveref}
\usepackage[absolute]{textpos}

\title{\LARGE \bf
Knowledge Distillation for Semantic Segmentation: A Label Space Unification Approach
}

\author{Anton Backhaus, Thorsten Luettel and Mirko Maehlisch*% <-this % stops a space
\thanks{This research paper is funded by dtec.bw – Digitalization and Technology Research Center of the Bundeswehr [project MORE] which we gratefully acknowledge. dtec.bw is funded by the European Union – NextGenerationEU.}% <-this % stops a space 
\thanks{*All authors are with the Chair of Machine Perception for Autonomous Driving, Department of Aerospace Engineering, University of the Bundeswehr Munich, Germany. Contact author email: {\tt\small anton.backhaus@unibw.de}}%
}

\begin{document}
% \copyrightstatement

\bstctlcite{bibcontrol_etal2}

\maketitle
\thispagestyle{empty}
\pagestyle{empty}

%%%%%%%%%%%%%%%%%%%%%%%%%%%%%%%%%%%%%%%%%%%%%%%%%%%%%%%%%%%%%%%%%%%%%%%%%%%%%%%%
\begin{strip}
\centering
\vspace{-2cm}\\
% Include the four images
\includegraphics[
    trim={22cm 0 15cm 0},
    clip,
    width=0.248\textwidth
]{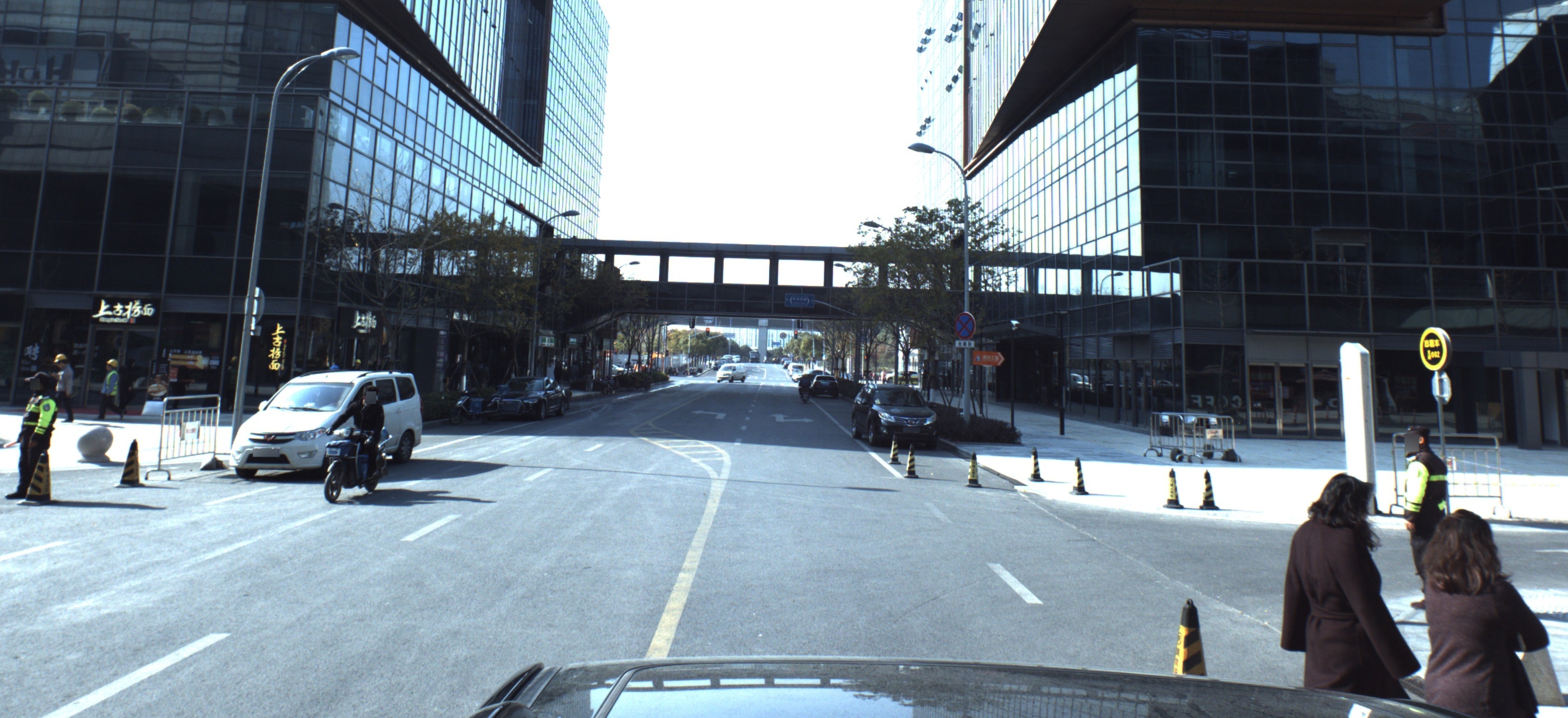}%
\hfill
\includegraphics[
    trim={22cm 0 15cm 0},
    clip,
    width=0.248\textwidth
]{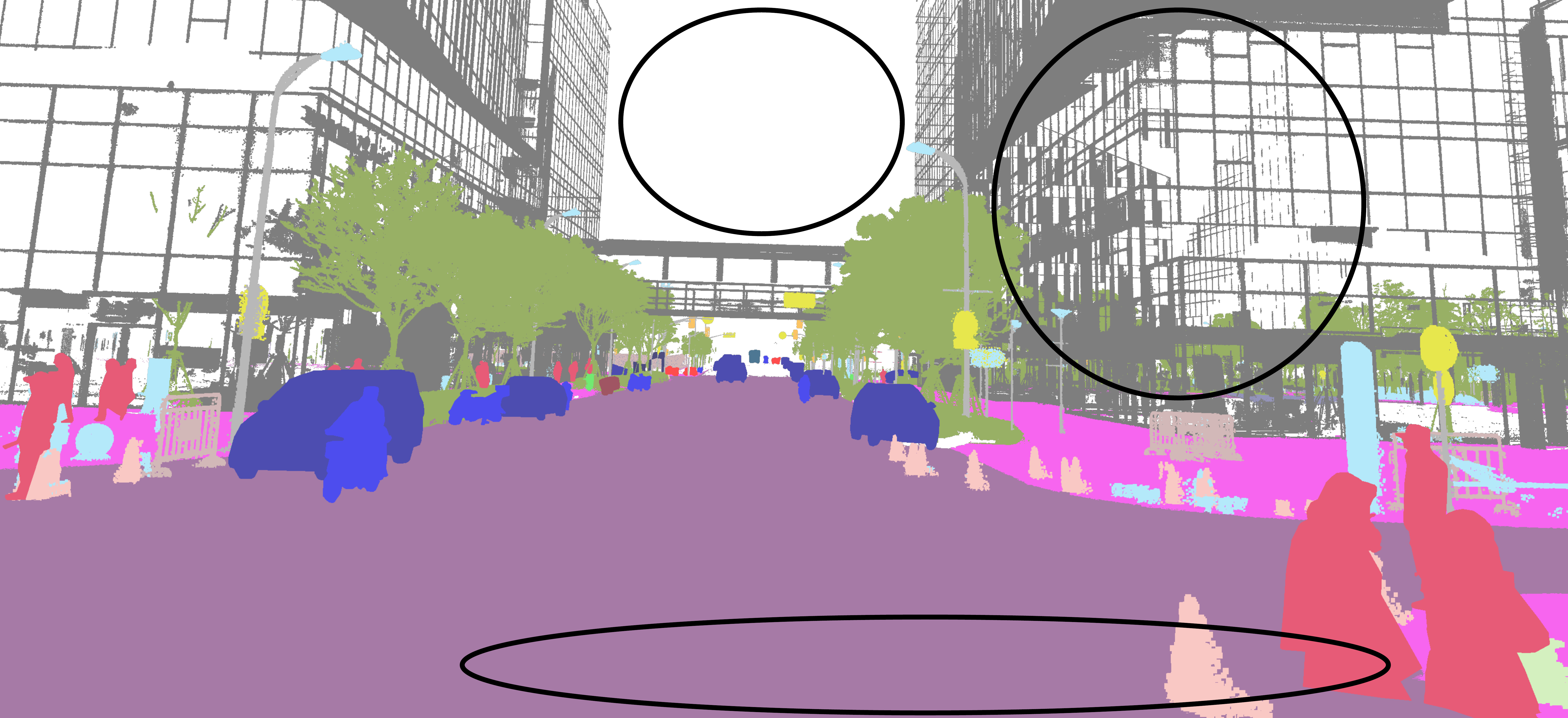}%
\hfill
\includegraphics[
    trim={22cm 0 15cm 0},
    clip,
    width=0.248\textwidth
]{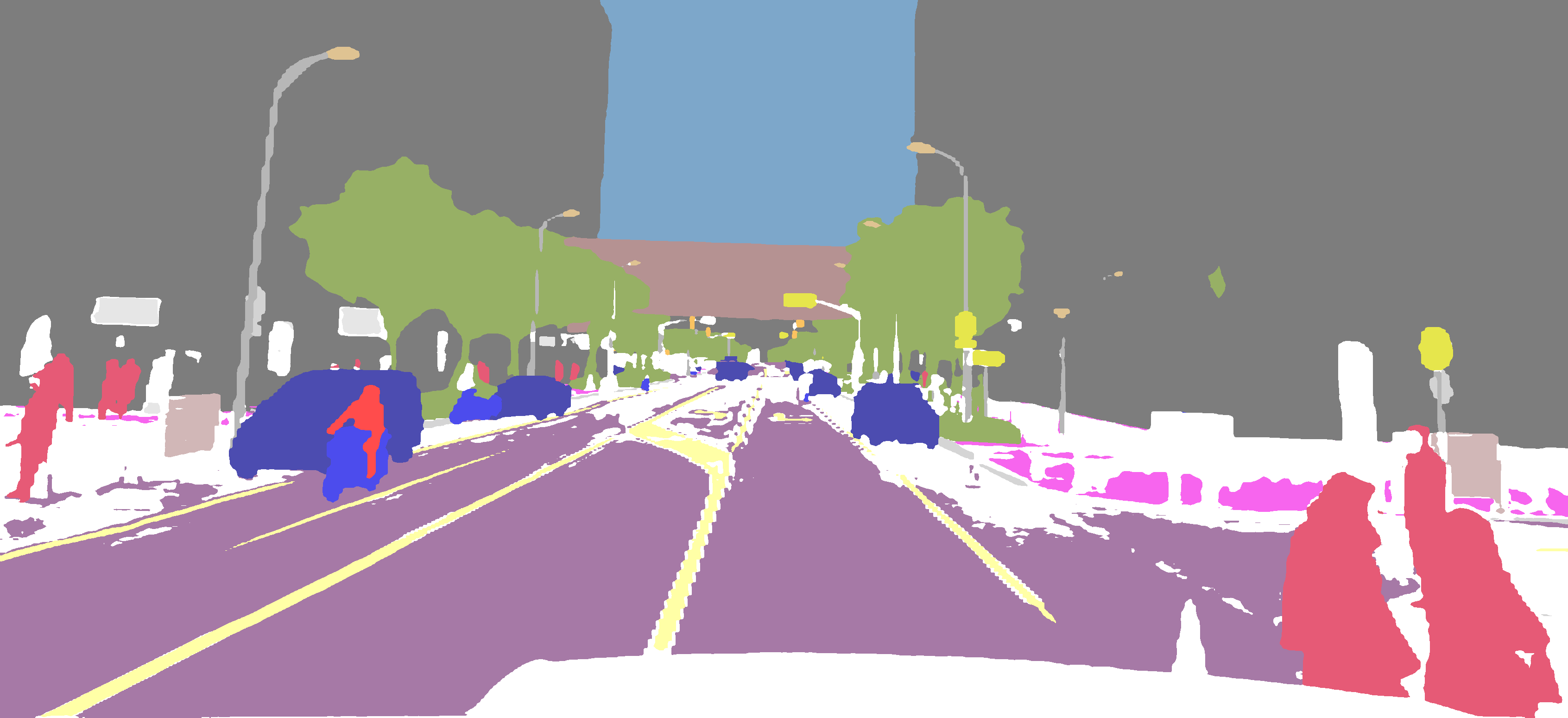}%
\hfill
\includegraphics[
    trim={22cm 0 15cm 0},
    clip,
    width=0.248\textwidth
]{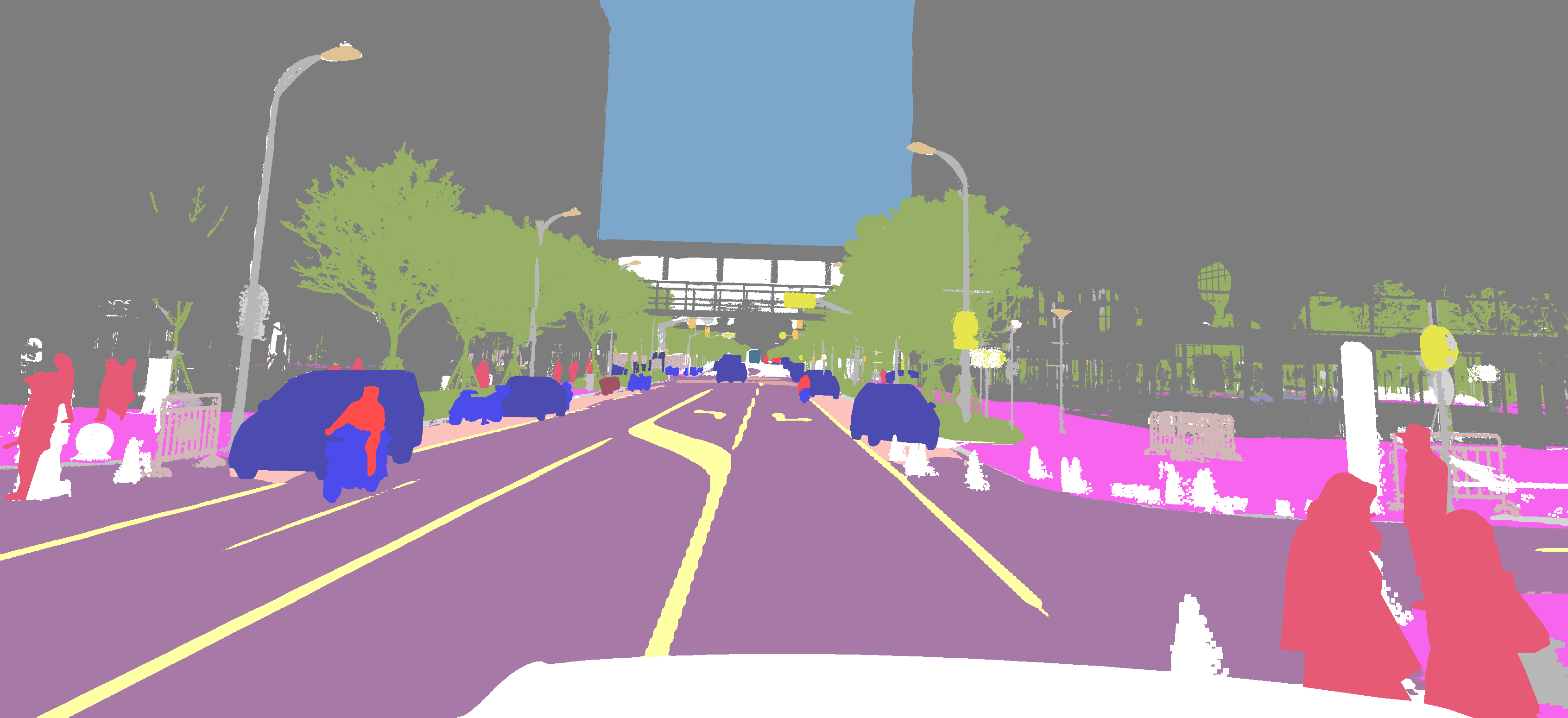}%

\vspace{2mm}

\begin{minipage}[t]{0.24\textwidth}
\centering
(a) RGB Image
\end{minipage}
\hfill
\begin{minipage}[t]{0.24\textwidth}
\centering
(b) Ground Truth
\end{minipage}
\hfill
\begin{minipage}[t]{0.24\textwidth}
\centering
(c) Prediction
\end{minipage}
\hfill
\begin{minipage}[t]{0.24\textwidth}
\centering
(d) Pseudo-Label
\end{minipage}

\vspace{0.5mm} % Adds a small vertical space before the caption

% Add the caption
\captionof{figure}{Pseudo-label creation for an ApolloScape \cite{bib:apolloscape} image: Using the ground truth (b), taxonomy mapping, and teacher model prediction (c), a pseudo-ground truth (d) is generated in the source label space. Systematic errors caused by the LiDAR-based semi-automated labeling like the missing labels for sky, building, and ego-vehicle (circled) are also fixed.}
\label{fig:title_image}
\end{strip}

\begin{abstract}

An increasing number of datasets sharing similar domains for semantic segmentation have been published over the past few years.
But despite the growing amount of overall data, it is still difficult to train bigger and better models due to inconsistency in taxonomy and/or labeling policies of different datasets.
To this end, we propose a knowledge distillation approach that also serves as a label space unification method for semantic segmentation.
In short, a teacher model is trained on a source dataset with a given taxonomy, then used to pseudo-label additional data for which ground truth labels of a related label space exist.
By mapping the related taxonomies to the source taxonomy, we create constraints within which the model can predict pseudo-labels.
Using the improved pseudo-labels we train student models that consistently outperform their teachers in two challenging domains, namely urban and off-road driving.
Our ground truth-corrected pseudo-labels span over 12 and 7 public datasets with 388.230 and 18.558 images for the urban and off-road domains, respectively, creating the largest compound datasets for autonomous driving to date.

\end{abstract}

%\renewcommand{\figurename}{Abbildung}
%\renewcommand{\tablename}{Tabelle}

%%%%%%%%%%%%%%%%%%%%%%%%%%%%%%%%%%%%%%%%%%%%%%%%%%%%%%%%%%%%%%%%%%%%%%%%%%%%%%%%
\section{Introduction}
% TODO: use the word model-agnostic somewhere
Transformers have become a popular choice of architecture for many vision-related tasks because of how well their performance scales with network and dataset size \cite{bib:attention-is-all-you-need}.
This is, however, both a strength and a weakness: When insufficient labeled data exists, transformers, particularly larger ones, will overfit and underperform.
At the same time, increasing amounts of labeled data are being published.
Unfortunately, in the field of semantic segmentation, it is uncommon for different datasets to share identical taxonomies and labeling policies, even if the target domain is the same, making it difficult to combine data.

The main issues typically arise from overlapping classes and differences in class granularity across datasets.
Classes "overlap" when they are inconsistently grouped with other classes across datasets.
For example, A2D2 \cite{bib:a2d2} merges buses and heavy duty trucks, BDD100k \cite{bib:bdd100k} merges heavy duty trucks and pickups, and in Mapillary Vistas \cite{bib:mapillary-vistas} pickup trucks and cars are merged.
Moreover, class granularity varies particularly in off-road datasets:
The ground can, for example, be divided into road/grass \cite{bib:freiburg-forest}, rough trail/smooth trail/traversable grass \cite{bib:yamaha}, or even 11 different ground classes in the case of the GOOSE dataset \cite{bib:goose-dataset}.
Furthermore, even where taxonomies \textit{are} consistent, labeling errors may cause a degradation in accuracy for certain classes.
For instance, ApolloScape's LiDAR-assisted semi-automatic labeling method causes buildings to be partially assigned a void label due to reflections on windows or the sky to be sometimes unlabeled due to missing LiDAR points (see \cref{fig:title_image} (b)).

To address these problems, we were inspired by Naive Student \cite{bib:naive-student}, originally a method for knowledge distillation.
Our proposed approach unifies data from similar domains with conflicting semantic segmentation labels using a teacher model and simultaneously trains better student models.
Here, the role of the teacher model that we pre-train on a source dataset is to predict pseudo-labels for images from similar auxiliary datasets using a source taxonomy.
The teacher's predictions are constrained by a pre-defined ontology mapping between source and auxiliary taxonomy.
For instance, the ApolloScape \cite{bib:apolloscape} dataset does not distinguish between roads, markings, and parking areas like Mapillary-Vistas \cite{bib:mapillary-vistas}.
When performing inference on an ApolloScape image, we simply map all pixels predicted as road, markings, parking area, etc., to pixels labeled as road in the ground truth.

Code, weights, and ontology mappings can be found at \href{https://github.com/UniBwTAS/data-priors}{https://github.com/UniBwTAS/data-priors}.

% - training on multiple datasets has the positive effect of not only increasing performance on the target dataset but also increasing generalization capability, i.e. datasets where there is a domain gap

\section{Related Work}

\subsection{Knowledge Distillation}
Knowledge distillation is a technique in which a student model is usually trained by mimicking the outputs of a more powerful teacher model.
However, in a semi-supervised learning context, knowledge distillation can also be used to train same-size or larger student models that outperform their teachers by pseudo-labeling large amounts of unlabeled data \cite{bib:billion-scale, bib:naive-student, bib:noisy-student, bib:mean-teacher, bib:pp-liteseg, bib:meta-pseudo-labels, bib:data-distillation}.
Some studies even show that pre-training a model on its own pseudo-labels can produce better models than those pre-trained on another hand-labeled dataset \cite{bib:rethinking-pretraining, bib:depth-anything}.

In Naive Student \cite{bib:naive-student}, for example, the authors use a Cityscapes-trained model to pseudo-label abundant unlabeled images of the same dataset with semantic, instance, and panoptic segmentation labels.
The newly initiated student model then trains on all data (pseudo and ground truth).
The entire process is iterated through multiple times, whereby for each iteration, the student model becomes the teacher model for the next iteration.
In a similar work, Noisy Student \cite{bib:noisy-student}, a classification model, is used to generate soft pseudo-labels for 300M images.
During the subsequent student training, noise is injected into the model, improving generalizability and overall classification accuracy.

The above works focus on utilizing large amounts of unlabeled data.
Another approach is to use weakly annotated data, e.g. in the form of object bounding boxes \cite{bib:weakly-and-semi-supervised-learning, bib:boosting-ssss, bib:bbam}, rough segmentation maps \cite{bib:urban-scene-coarse-annotation} or image-level annotations \cite{bib:weakly-and-semi-supervised-learning, bib:boosting-ssss, bib:image-level-weak-supervision}.
Our work fits somewhere in between these two paradigms, since it uses pseudo-labels that were enhanced by dataset priors (also a form of weak supervision) but is otherwise fully compatible with unlabeled data.

\subsection{Label Space Unification}

Label Space Unification for semantic segmentation is typically approached from three angles: at the model architecture level \cite{bib:automated-label-unification, bib:multi-head-semseg, bib:cross-dataset-learning, bib:universal-ssss, bib:heterogeneous-street-datasets, bib:multi-task-domain-learning}, the loss level \cite{bib:universal-image-concepts, bib:partial-label-losses, bib:multi-domain-semseg, bib:heterogeneous-street-datasets, bib:multi-task-domain-learning}, and the data level \cite{bib:mseg, bib:unifying-off-road}.

Modifying the architecture to accommodate different label spaces typically involves adding prediction heads for each taxonomy \cite{bib:heterogeneous-street-datasets, bib:multi-head-semseg, bib:universal-ssss, bib:multi-task-domain-learning}, or extra layers that learn the taxonomy between the label spaces \cite{bib:automated-label-unification}.
However, besides increasing the overall model size and complexity, elaborate model architectures have limited practical use and can be incompatible with existing architectures.

Moreover, past works have proposed methods whereby only prediction errors within the dataset-specific label space are penalized by the loss function, leaving the model architecture largely unchanged:
Bevandi\'{c} et al. \cite{bib:universal-image-concepts, bib:multi-domain-semseg} introduced a variation of the negative log-likelihood loss that is designed to handle partial labels, i.e., labels that represent a set of possible classes in a hierarchical taxonomy.
They do this by aggregating the probabilities of universal classes corresponding to each dataset-specific label and computing the log-sum over these probabilities.
Similarly, in \cite{bib:heterogeneous-datasets} the softmax output is re-mapped from the fine-granular universal label space to a dataset-specific one before applying the standard cross-entropy loss.
Fourure et al. \cite{bib:multi-task-domain-learning} avoid the the label space unification step by letting the model predict over all classes of all datasets within the softmax layer.
Via their proposed "selective cross-entropy" loss function, only logits belonging to the dataset of the input image have a non-zero gradient in the backpropagation step.
However, such methods of naive label space concatenation bring with them significant overhead in train and prediction time since the dimension of the softmax output grows with each class.
Modifying only the loss function has the advantage of reduced data preparation.
However, this approach often requires heavy modifications of the training pipeline and not all models may be compatible with the proposed loss function.

Finally, a straightforward but time-consuming approach to dataset unification is via manipulating the data itself.
MSeg \cite{bib:mseg} unifies 7 datasets under a fine-granular taxonomy by manual labeling.
In a similar work \cite{bib:unifying-off-road}, the authors unify several off-road datasets by using an ontology-based framework that maps differing label schemes to a shared taxonomy. 
Both approaches lack an automated labeling scheme, thus necessitating manual re-labeling to a source taxonomy or moving classes to the highest level hierarchy that is shared across all datasets, resulting in a loss of granularity.

Our approach, due to its simplicity, does not require any modification to architecture or loss function and can be easily built on top of existing training pipelines.

\section{Methodology}

\begin{figure*}[!ht]
    \centering
    \includegraphics[width=\textwidth]{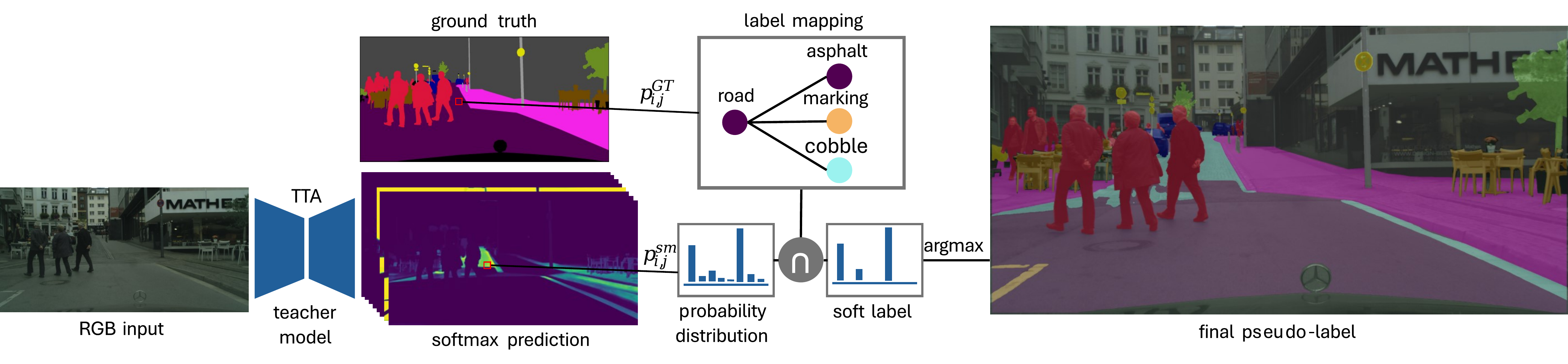} 
    \caption{Pseudo-label generation process: An RGB image is fed into a teacher network. The resulting softmax output is first refined using test-time augmentation (TTA), then using the ground truth. Here, pixel $p_{ij}$ is \texttt{cityscapes:road}. Additionally, using the label mapping, the output at $p_{ij}$ is constrained to \texttt{GOOSE:asphalt} (\textcolor[rgb]{0.502, 0.251, 0.502}{\rule{1.5ex}{1.5ex}}), \texttt{marking} (\textcolor[rgb]{0.9647, 0.7059, 0.3922}{\rule{1.5ex}{1.5ex}}) or \texttt{cobble} (\textcolor[rgb]{0.6078, 0.949, 0.9373}{\rule{1.5ex}{1.5ex}}). The final hard pseudo-label is depicted on the right.}
    \label{fig:method}
\end{figure*}

The pseudo-labeling process is illustrated in \cref{fig:method} exemplary for a Cityscapes image using the GOOSE taxonomy.
The elaboration below follows terminology defined in \cite{bib:multi-domain-semseg}.

\textbf{Teacher Training}  
A teacher model \(\theta_t\) is trained on a source dataset \(\mathbb{D}_s = \{(x^s, y^s)\}\), where \( x^s \) are input images, and \( y^s \) are their corresponding dense labels within a taxonomy \(\mathbb{S}_s\).
The objective minimizes a segmentation loss \(\mathcal{L}_{\text{seg}}\) over all samples in \(\mathbb{D}_s\):
\begin{equation*}
\theta_t^* = \arg \min_{\theta_t} \frac{1}{N} \sum_{i=1}^N \mathcal{L}_{\text{seg}}(y_i^s, f(x_i^s; \theta_t)),
%\tag{1} 
\end{equation*}
where \(f(x; \theta_t)\) represents the teacher model's predictions.

\textbf{Ontology Mapping}  
We define the set of all extra datasets \(\mathbb{D}_e\) as such that they do not have identical taxonomies to the source dataset but ideally intersect:
\[
\mathbb{S}_i \neq \mathbb{S}_s \land \mathbb{S}_i \cap \mathbb{S}_s \neq \emptyset, \forall \mathbb{D}_i \in \mathbb{D}_e.
\]
Manually constructed ontologies \(\mathcal{O}_{e \to s}\) map labels between \(\mathbb{S}_e\) and \(\mathbb{S}_s\).
The mapping defines valid class relationships, e.g.
\[
\texttt{CS:road} \to \{\texttt{GOOSE:asphalt}, \texttt{marking}, \texttt{cobble}\},
\]
source classes that do not appear in the extra dataset, e.g.
\[
\texttt{MV:ground-animal} \notin \mathbb{D}_{apolloscape},
\]
and can account for systematic labeling errors as shown in \cref{fig:title_image} (b):
\[
\texttt{MV:building} \to \{\texttt{ApolloScape:building}, \texttt{void}\}.
\]

Excluding classes from an ontology or mapping rare ones that do not appear in the extra taxonomy can be time-consuming because it often necessitates looking through a large subset of the extra dataset.
Ground animals, for example, are rarely explicitly labeled and often conflated with the \texttt{void} class and are sometimes not contained at all in some autonomous driving datasets, but to confirm this would involve checking a large subset of images.
However, it is viable for smaller datasets or such that have consecutive frames taken in short time and distance from one another.
ApolloScape, for example, has roughly 140.000 frames.
These contain a left and right image, of which it is sufficient to look at every 20th frame without the scene changing too much.
The number of completely unique frames in this case is thus easily reduced 3.500.
Furthermore, for creating an ontology mapping, checking a subset of frames is usually sufficient.

\textbf{Prediction Refinement:}
The teacher model predicts soft labels for \(\mathbb{D}_e\) using test-time augmentations \(\text{Aug}(\cdot)\):
\[
\tilde{y}_i^e = \frac{1}{|\mathcal{A}|} \sum_{a \in \mathcal{A}} f(\text{Aug}_a(x^e); \theta_t^*),
%\tag{2}
\]
where \(\mathcal{A}\) denotes the set of augmentations.
We apply a set of standard augmentations, including horizontal flipping and multi-scale inputs at scales of 0.5, 0.75, 1, 1.25, 1.5, 1.75, 2.
The superposed soft predictions are finally refined using the ontology.
% \mathbb{O}_{s \to e}(p_{ij}^{GT}
The refinement step sets all softmax scores for implausible predictions to zero for each pixel \( p_{ij} \) of an input image $x^e$:
\[
\tilde{p}_{ij}(c) =
\begin{cases}
    p_{ij}(c), & \text{if } c \in \mathcal{O}_{s \to e}(p_{ij}^{GT}) \\
    0, & \text{otherwise}
\end{cases}
\quad \forall p_{ij} \in x^e
\]

where:
\begin{itemize}
    \item \( p_{ij}^{GT} \): ground truth label for pixel \( p_{ij} \) in the source dataset
    \item \( \mathcal{O}_{s \to e}(p_{ij}^{GT}) \): ontology-mapped set of possible labels in the extra dataset for that specific ground truth class
    \item \( p_{ij}(c) \): the softmax score for class \( c \)
    \item \( \tilde{p}_{ij}(c) \): final softmax score for a given class
\end{itemize}
The ontology-corrected soft predictions are finally turned into hard pseudo-labels via $\arg\max$ to save disk space on the one hand and to make it generally compatible for most semantic segmentation models on the other hand.
Some state-of-the-art models like Mask2Former \cite{bib:mask2former} and OneFormer \cite{bib:oneformer} use Hungarian Matching in their loss function which is incompatible with soft labels.
However, there is evidence suggesting slightly better performance using soft labels \cite{bib:noisy-student}.

\textbf{Student Training:}
Once all extra data have refined pseudo-labels contained in the source label space, we train a student model on the unification of source dataset and extra datasets.
During training, the intersection over union (IoU) is measured on the validation split of the source dataset and, if applicable, on WildDash \cite{bib:wilddash}.
When there is no improvement after 15 epochs, the best checkpoint is fine-tuned only on the source dataset with a \( 0.1 \times \) learning rate.
The fully trained student model can optionally become the teacher model in another iteration of training.

% The conventional approach is to pre-train a model on a large, diverse dataset such as COCO, then fine-tune it with only the final prediction layer adjusted to match the number of classes.
% - construct a unified label space
% - A downside to this method is that although more data has a positive effect on accuracy, it also increases the training time for each new student model
% - even if the sample image is strongly out of distribution (show coco) the model manages to produce a satisfactory pseudo label when guided by priors
% - light augmentation

% show example ontology of GOOSE and cityscapes; write a short description of how many classes the models have

\section{Experiments and Results}

To evaluate the efficacy of our method we conduct a series of experiments whereby we vary the size of the model, the size of the source dataset that the teacher model is initially trained on, the domain (on/off-road) and training with and without priors. 
Using no priors is equivalent to Naive Student \cite{bib:naive-student}.
The evaluation is done with two different sizes of Mask2Former: Large and Base.
We do this to better control for over-fitting of large teacher models on smaller datasets in the initial training phase.
An overview of the used datasets can be found in \cref{tab:dataset_overview}.

\begin{table}[t]
    \centering
    \caption{List of urban (top) and off-road (bottom) datasets by number of frames. Datasets are considered contiguous when frames were taken in close spatio-temporal proximity.}
    \label{tab:dataset_overview}
    \small
    \begin{tabular}{l c c c}
        \toprule
        \textbf{Dataset} & \textbf{Frames} & \textbf{Rel. [\%]} & \textbf{Contiguous} \\
        \midrule
        ApolloScape \cite{bib:apolloscape}      & 131,286 & 34  & \checkmark \\
        NuImages \cite{bib:nuscenes}            & 83,724  & 22  & \checkmark \\
        Waymo \cite{bib:waymo-dataset}         & 75,680  & 19  & \checkmark \\
        A2D2 \cite{bib:a2d2}                    & 41,277  & 11  & \checkmark \\
        MV \cite{bib:mapillary-vistas}          & 20,000  & 5   &   \\
        IDD \cite{bib:idd}                      & 16,063  & 4   &   \\
        BDD \cite{bib:bdd100k}                  & 8,000   & 2   &   \\
        COCO* \cite{bib:coco}                   & 5,711   & 1   &   \\
        Cityscapes \cite{bib:cityscapes}        & 3,475   & 1   &   \\
        GOOSE* \cite{bib:goose-dataset}         & 2,172   & 1   & \checkmark \\
        CamVid \cite{bib:camvid}                & 469     & \textless1  & \checkmark \\
        Lanes \cite{bib:lanes}                  & 373     & \textless1  & \checkmark \\
        \midrule
        \textbf{Total urban}                    & 388,230 &    &   \\
        \midrule
        GOOSE \cite{bib:goose-dataset}          & 8,816   & 48  & \checkmark \\
        Rellis3D \cite{bib:rellis-3d}           & 4,285   & 23  & \checkmark \\
        Cityscapes \cite{bib:cityscapes}        & 3,475   & 19  &   \\
        YCOR \cite{bib:yamaha}                  & 1,076   & 6   &   \\
        TAS500 \cite{bib:tas500}                & 540     & 3   &   \\
        FreiburgForest \cite{bib:freiburg-forest} & 366   & 2   &   \\
        \midrule
        \textbf{Total off-road}                 & 18,558  &    &   \\
        \bottomrule
    \end{tabular}
\end{table}

\subsection{Urban Datasets}

To test our method we trained a model on 12 datasets \cite{bib:a2d2, bib:apolloscape, bib:bdd100k, bib:camvid, bib:idd, bib:mapillary-vistas, bib:waymo-dataset, bib:nuscenes, bib:coco, bib:goose-dataset, bib:cityscapes, bib:lanes}, including a subset of COCO and GOOSE depicting urban scenes.
Cityscapes \cite{bib:cityscapes} and Mapillary-Vistas \cite{bib:mapillary-vistas} are used as the source dataset for initial teacher training and defining the taxonomy.
We also use WildDash \cite{bib:wilddash} for evaluating robustness and generalizability but exclude it from training.
Cityscapes and Mapillary-Vistas consist of 3475 and 20.000 annotated publicly available images, respectively.
The additional data adds just under 400.000 training images.
% Train on xy size images with random crop
% naive student trains on 170.000 images
As a baseline, we also trained Mask2Former Large on Cityscapes following the Naive Student training protocol but using the unlabeled Cityscapes extra data instead of the above mentioned datasets.
All models were initialized with ADE20K \cite{bib:ade20k} pre-trained weights.

\begin{table}[b]
\centering
\scriptsize
\setlength{\tabcolsep}{4pt} % Reduce column spacing
\caption{Evaluation of Mask2Former (M2F) Large (L) and Base (B) on Cityscapes val split (19 classes), Mapillary-Vistas val split (63 classes) and WildDash for one and two iterations as well as our baseline M2F Naive Student (NS) trained on Cityscapes extra data.
We compare initial IoU in \% of the teacher (Init.) and post-training results (Post) of the student model with and without priors.
Best results on the Cityscapes taxonomy are boldfaced.}
\label{tab:urban-results}
\begin{tabular}{cccccccc}
\toprule
\textbf{Model} & \textbf{It.} & \textbf{Priors (ours)} & \multicolumn{2}{c}{\textbf{Cityscapes}} & \multicolumn{2}{c}{\textbf{WildDash}} \\
\cmidrule(lr){4-5} \cmidrule(lr){6-7}
 &  &  & Init. & Post (diff) & Init. & Post (diff) \\
\midrule
M2FL+NS & 1 &  & 73.6 & 77.8 (+4.2) & 56.9 & 50.2 (+6.7) \\
M2FL+NS & 2 &  & 77.8 & 73.9 (-3.9) & 50.2 & 48.2 (-2.0) \\
\arrayrulecolor{gray}
\hdashline
\arrayrulecolor{black}
M2FB & 1 &  & 73.6 & 73.0 (-0.6) & 54.5 & 51.9 (-2.6) \\
M2FB & 1 & \checkmark & 73.6 & 75.5 (+1.9) & 54.5 & 61.0 (+6.5) \\
M2FL & 1 &  & 73.4 & 76.8 (+3.4) & 56.9 & 56.5 (-0.4) \\
M2FL & 1 & \checkmark & 73.4 & \textbf{78.3 (+4.9)} & 56.9 & \textbf{66.5 (+9.6)} \\
M2FL & 2 & \checkmark & 78.3 & 76.7 (-1.6) & 66.5 & 66.2 (-0.3) \\
\midrule
\textbf{Model} & \textbf{It.} & \textbf{Priors (ours)} & \multicolumn{2}{c}{\textbf{Mapillary}} & \multicolumn{2}{c}{\textbf{WildDash}} \\
\midrule
M2FL & 1 & \checkmark & 52.9 & 52.7 (-0.2) & 34.2 & 38.8 (+4.6) \\
\bottomrule
\end{tabular}
\end{table}

Table \ref{tab:urban-results} shows a comparison of trained Mask2Former models rated by mean Intersection over Union (mIoU).
Our key findings are as follows:

\textbf{1. The improvement for the large model is greater than that of the base model.}
Although initially both sizes of the teacher model score roughly the same mIoU on Cityscapes, the base model improves only by 1.9\% versus the large model with 4.9\%.
This confirms the notion that larger models profit more from larger datasets.

\textbf{2. Training on slightly domain shifted but partially labeled data is more effective than training on same-domain fully pseudo-labeled data.}
This is demonstrated by the fact that the baseline, Naive Student, which was trained on Cityscapes pseudo-labeled extra data achieved a slightly lower IoU score (77.8\%) than our best model (78.3\%) and a much lower score on WildDash (50.2\% vs 66.5\%) despite originating from the same teacher model.
However, an important caveat here is that our partially labeled compound dataset contains roughly twice as many images as the Cityscapes extra data.

\textbf{3. Using dataset priors is always better than not.}
If labels exist, albeit with a different label space, it is always better to use the information they provide.
This is particularly true for datasets with a domain shift because inference on these data will often be inaccurate if not guided by partial labels.
Training with inaccurate pseudo-labels can even reduce performance of the student relative to the teacher model (see M2FB).

\textbf{4. Further iterations do not improve accuracy.}
This was the case both for our method as well as the reproduced Naive Student.
This was rather surprising as it goes contrary to the findings of Chen et al. \cite{bib:naive-student} who continuously improved their student model by using it as the teacher in the next iteration of pseudo-labels.
The key difference might be Mask2Former.
Mask2Former uses hard mining in its mask loss, i.e. it samples points in the prediction that have a high uncertainty.
This could lead to overly sampling wrong pseudo-labels effectively causing a degradation of performance across iterations.
% We discuss this problem in further detail in \cref{chapter:confirmation-bias}.

\textbf{5. The performance improvement is large and consistent on WildDash but not consistent on the source dataset.}
Since the models are trained on multiple diverse datasets, the improvement on WildDash is unsurprising; WildDash is a dataset specifically made to measure robustness and generalizability.
We did not, however, see any improvement on Mapillary-Vistas as the source dataset.
Possible reasons for this are imprecise ontology mappings due to the large number of classes (66) or that Mapillary-Vistas being an already large and diverse dataset has less potential for improvement than Cityscapes.

% TODO: important: compare no priors areas to priors areas --> how did priors guidance affect the prevalence of certain classes?

\subsection{Off-Road Datasets}

\begin{table}[b]
\centering
\scriptsize
\setlength{\tabcolsep}{4pt}
\caption{Evaluation of Mask2Former (M2F) Base (B) and two iterations of Large (L) on source GOOSE. We compare initial mean intersection over union (Init.) in \% and post-training results (Post) with and without priors. The best results are boldfaced.}
\label{tab:goose-results}
\begin{tabular}{cccccc}
\toprule
\textbf{Model} & \textbf{It.} & \textbf{Priors (ours)} & \multicolumn{2}{c}{\textbf{GOOSE}} \\
\cmidrule(lr){4-5}
 &  &  & Init. & Post (diff) \\
\midrule
M2FB & 1 & \checkmark & 61.3 & 64.4 (+3.1) \\
M2FL & 1 &  & 64.2 & 64.4 (+0.2) \\
M2FL & 1 & \checkmark & 64.2 & \textbf{67.9 (+3.7)} \\
M2FL & 2 & \checkmark & 67.9 & 67.3 (-0.6) \\
\bottomrule
\end{tabular}
\end{table}

\newlength{\gridColumnSpace}
\setlength{\gridColumnSpace}{0.24\textwidth}

\begin{figure*}[!t]
    \centering
    
    % Reduce horizontal spacing between columns
    \setlength{\tabcolsep}{1pt}
    % Slightly reduce vertical spacing between rows
    \renewcommand{\arraystretch}{0.85}
    
    \begin{tabular}{@{\hspace{0pt}}c@{\hspace{2pt}} c c c c}
        %-------------------------------------
        % Header row: column captions
        %   - The initial & shifts us into the second column
        %-------------------------------------
        &
        RGB Image &
        Ground Truth &
        Prediction &
        Pseudo-Label
        \\[1pt] % negative space to pull captions closer to the first row of images
        
        %-------------------------------------
        % Row 1: Cityscapes
        %-------------------------------------
        \raisebox{0.17\height}{\rotatebox{90}{Cityscapes}} &
        \begin{subfigure}[b]{\gridColumnSpace}
            \centering
            \includegraphics[width=\textwidth]{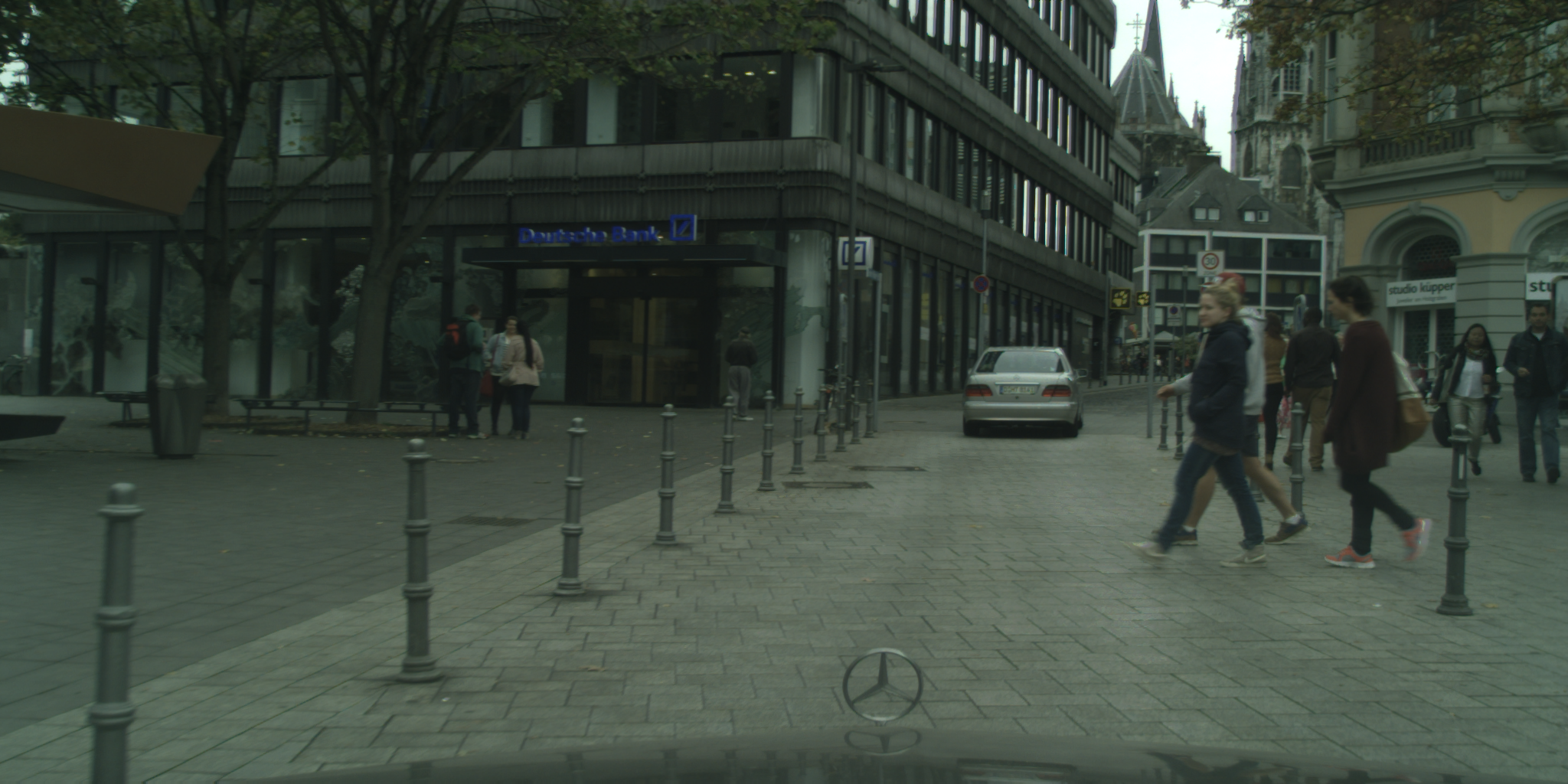}
        \end{subfigure} &
        \begin{subfigure}[b]{\gridColumnSpace}
            \centering
            \includegraphics[width=\textwidth]{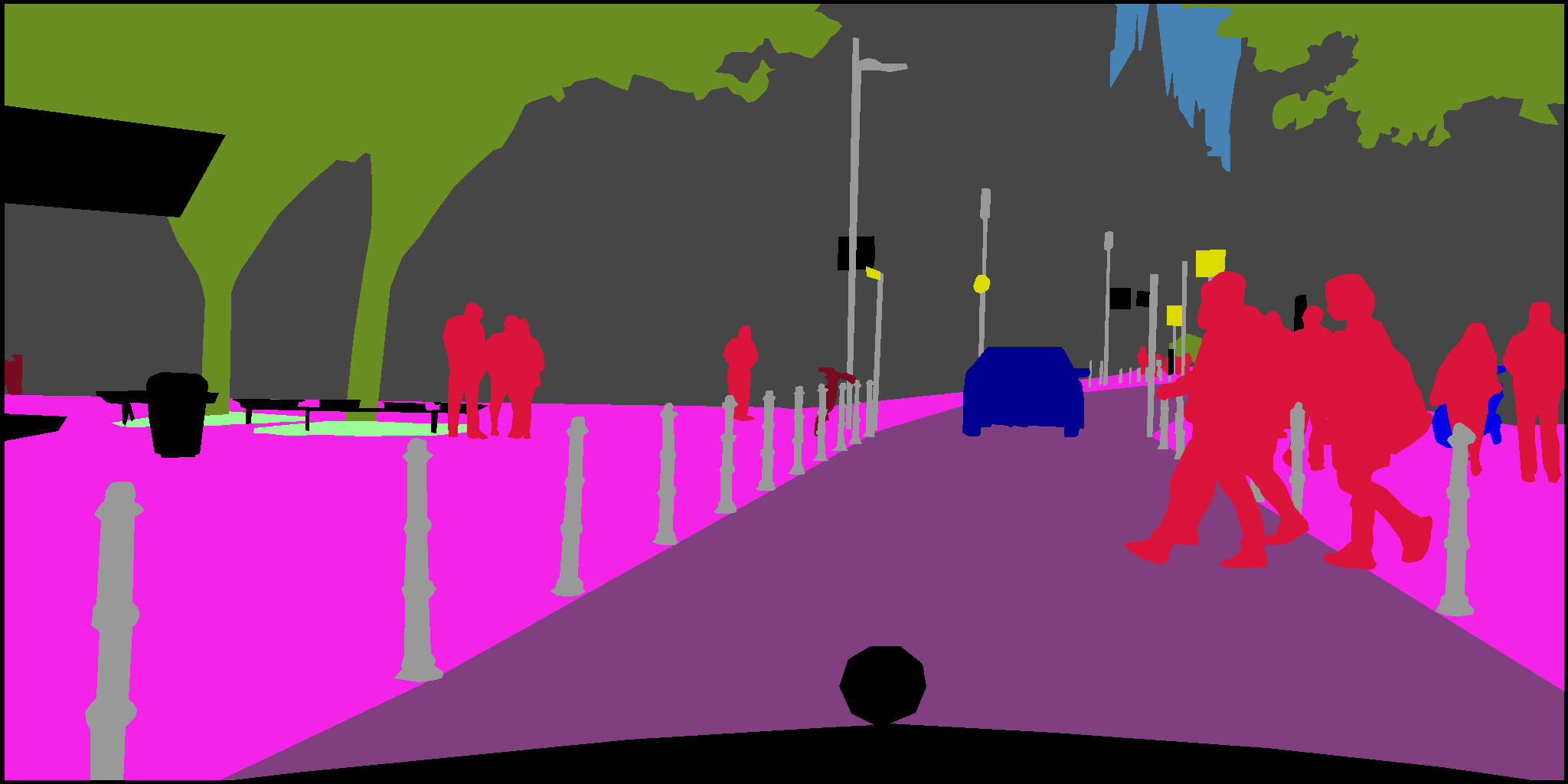}
        \end{subfigure} &
        \begin{subfigure}[b]{\gridColumnSpace}
            \centering
            \includegraphics[width=\textwidth]{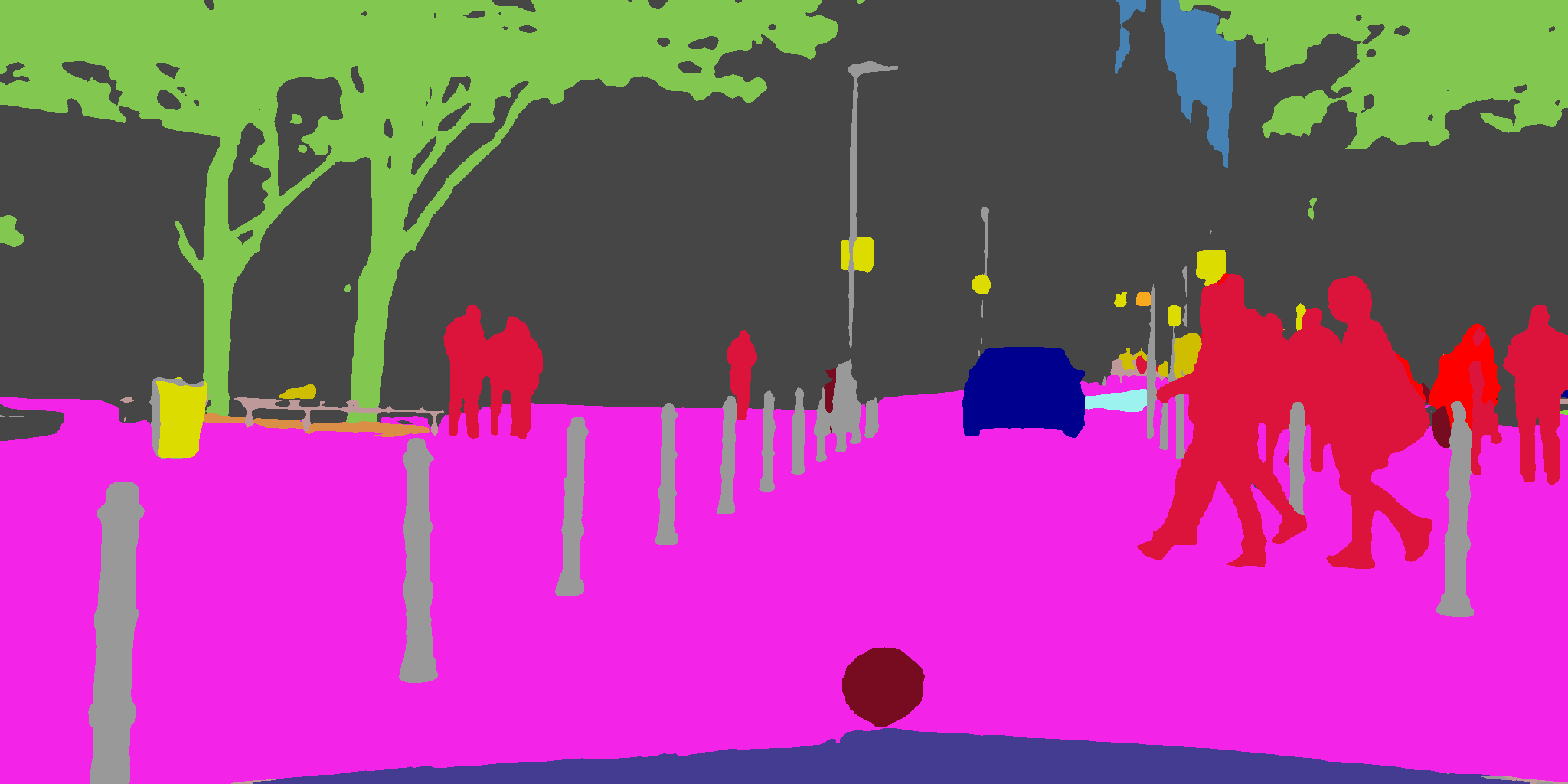}
        \end{subfigure} &
        \begin{subfigure}[b]{\gridColumnSpace}
            \centering
            \includegraphics[width=\textwidth]{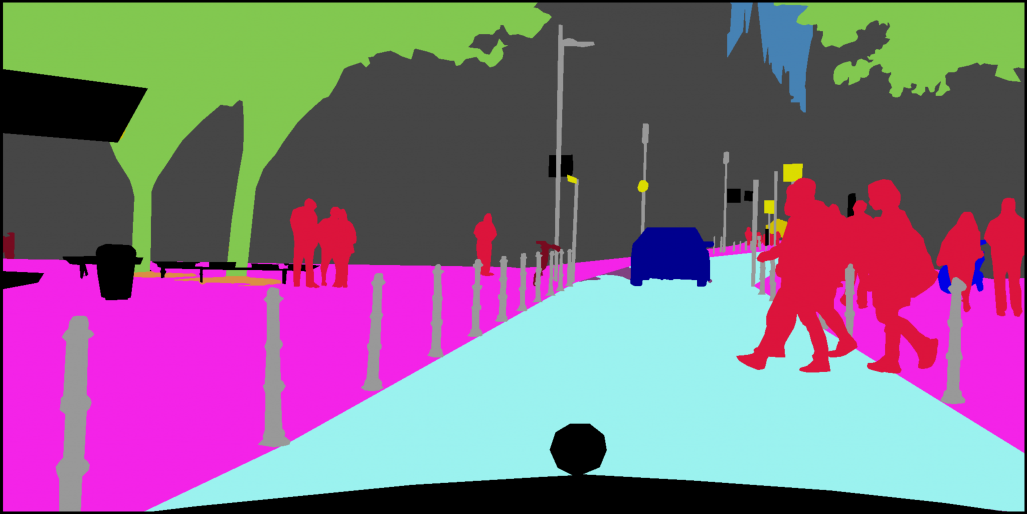}
        \end{subfigure}
        \\[-1pt] % reduce vertical space after row
        
        %-------------------------------------
        % Row 2: Rellis3D
        %-------------------------------------
        \raisebox{0.45\height}{\rotatebox{90}{Rellis3D}} &
        \begin{subfigure}[b]{\gridColumnSpace}
            \centering
            \includegraphics[width=\textwidth, trim=0 16 0 16, clip]{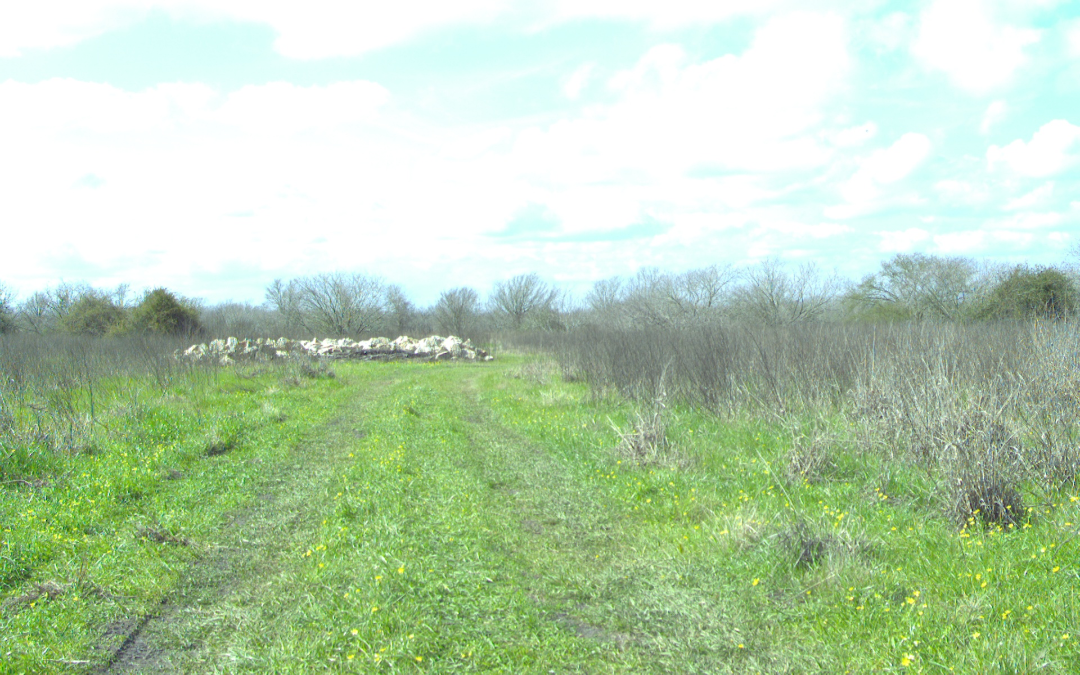}
        \end{subfigure} &
        \begin{subfigure}[b]{\gridColumnSpace}
            \centering
            \includegraphics[width=\textwidth, trim=0 16 0 16, clip]{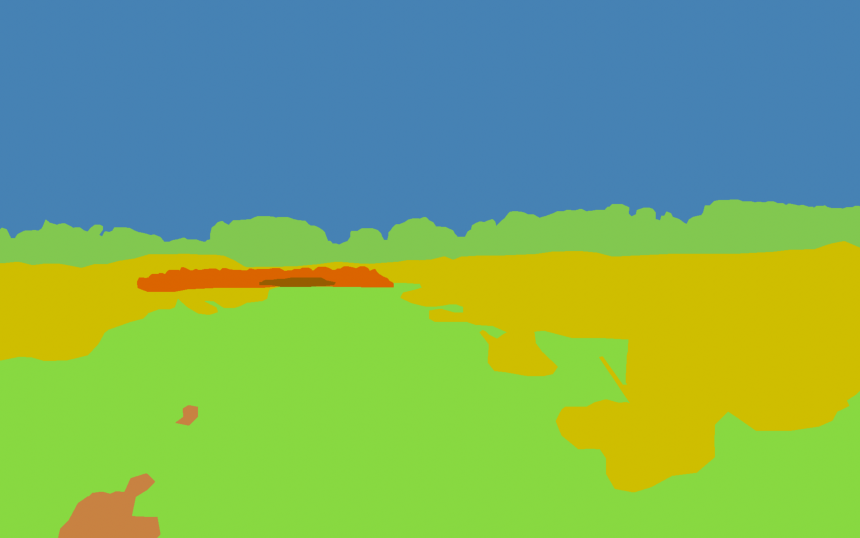}
        \end{subfigure} &
        \begin{subfigure}[b]{\gridColumnSpace}
            \centering
            \includegraphics[width=\textwidth, trim=0 13 0 13, clip]{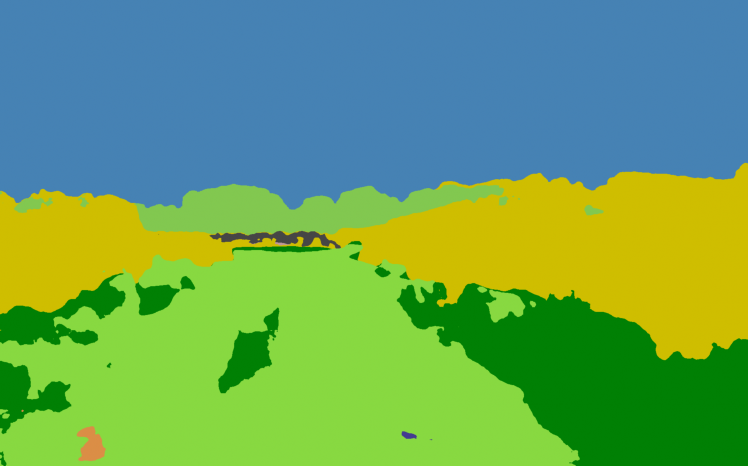}
        \end{subfigure} &
        \begin{subfigure}[b]{\gridColumnSpace}
            \centering
            \includegraphics[width=\textwidth, trim=0 18 0 18, clip]{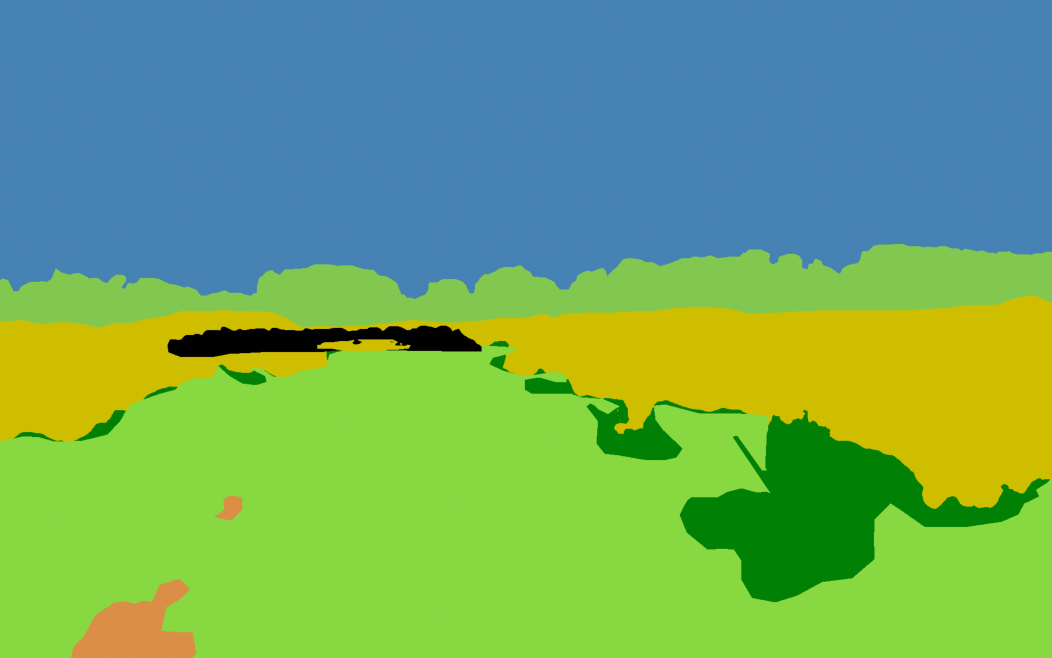}
        \end{subfigure}
        \\[-1pt]
        
        %-------------------------------------
        % Row 3: TAS500
        %-------------------------------------
        \raisebox{0.18\height}{\rotatebox{90}{TAS500}} &
        \begin{subfigure}[b]{\gridColumnSpace}
            \centering
            \includegraphics[
                trim={8cm 0 8cm 0},
                clip,
                width=\textwidth
            ]{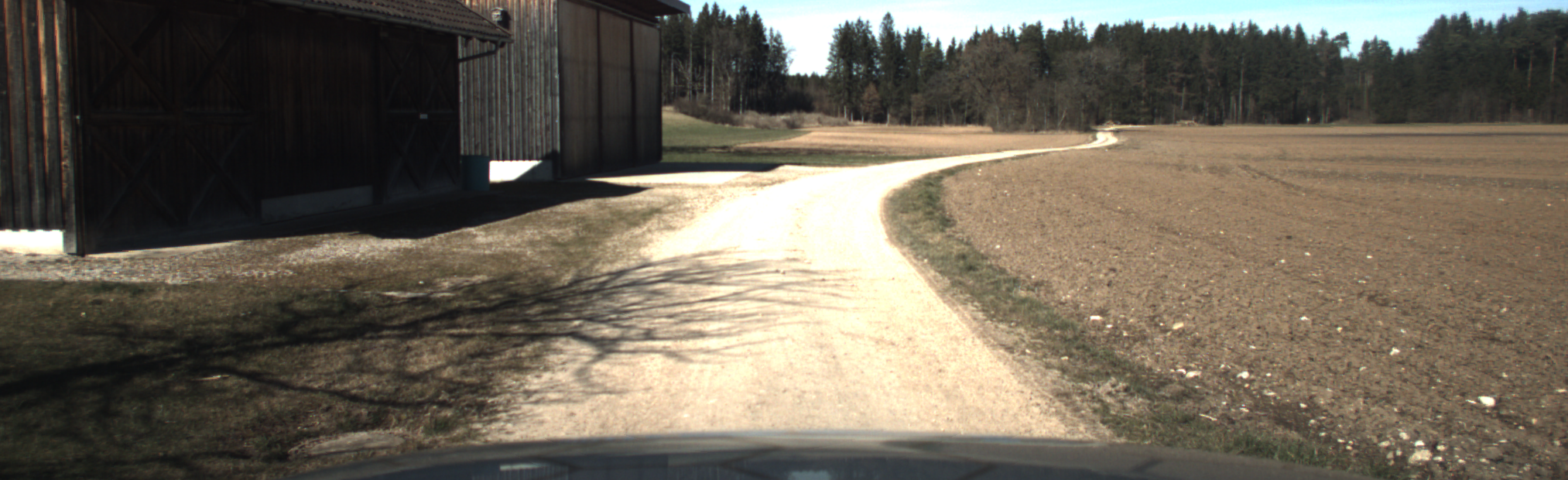}
        \end{subfigure} &
        \begin{subfigure}[b]{\gridColumnSpace}
            \centering
            \includegraphics[
                trim={8cm 0 8cm 0},
                clip,
                width=\textwidth
            ]{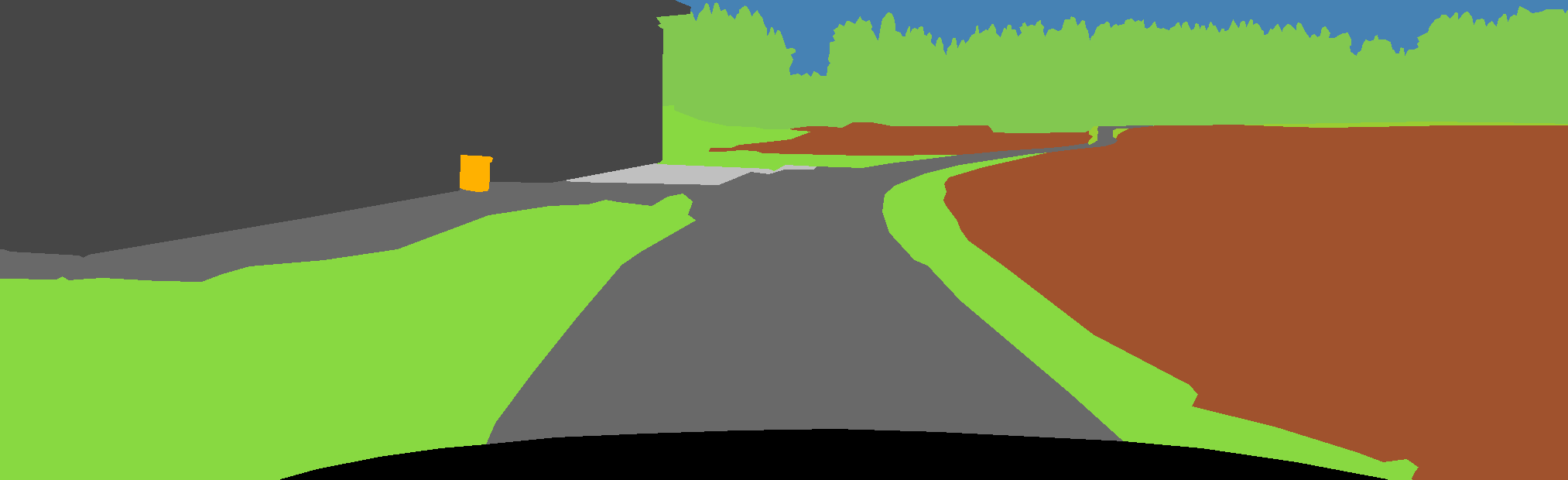}
        \end{subfigure} &
        \begin{subfigure}[b]{\gridColumnSpace}
            \centering
            \includegraphics[
                trim={8cm 0 8cm 0},
                clip,
                width=\textwidth
            ]{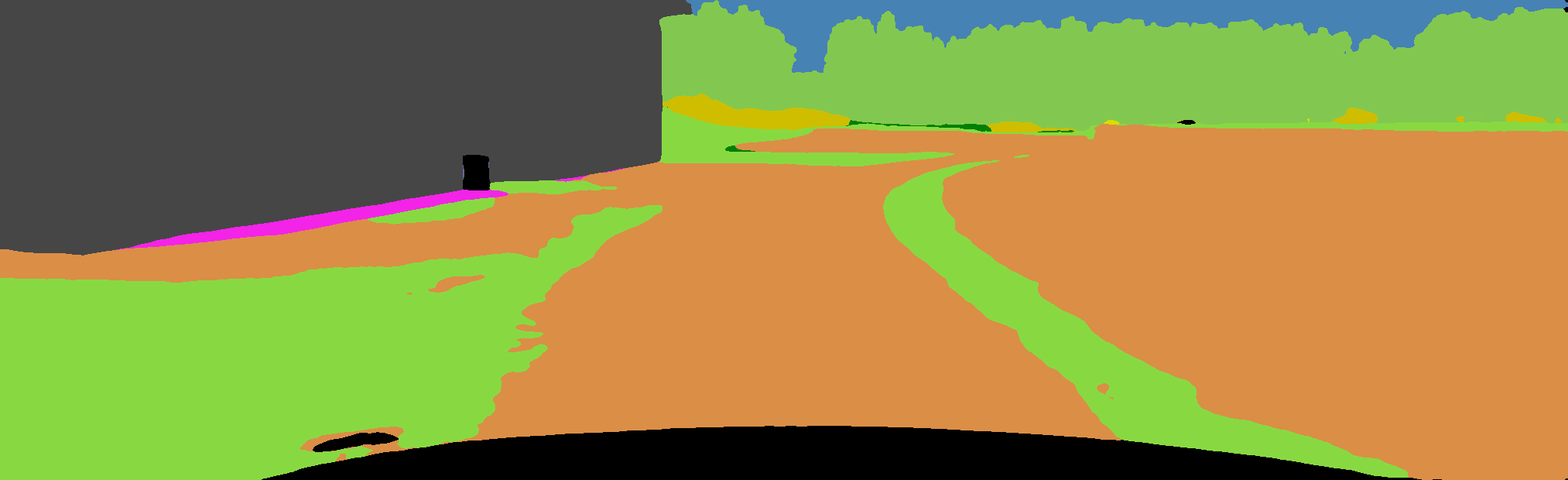}
        \end{subfigure} &
        \begin{subfigure}[b]{\gridColumnSpace}
            \centering
            \includegraphics[
                trim={8cm 0 8cm 0},
                clip,
                width=\textwidth
            ]{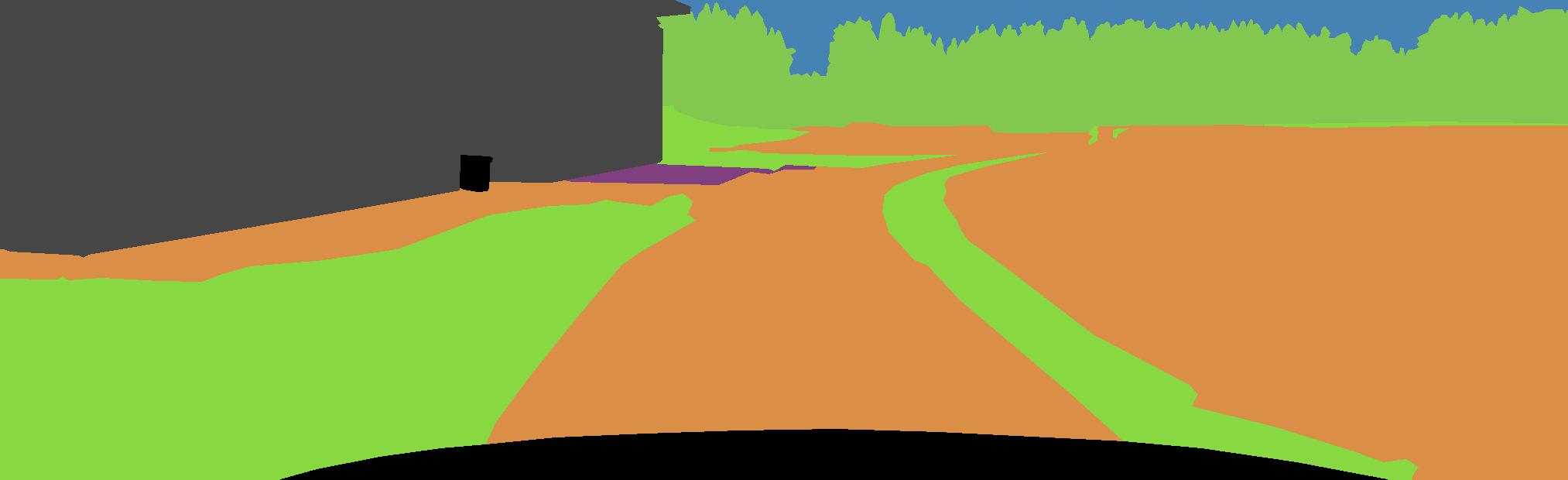}
        \end{subfigure}
        \\[-1pt]
        
        %-------------------------------------
        % Row 4: YCOR
        %-------------------------------------
        \raisebox{0.65\height}{\rotatebox{90}{YCOR}} &
        \begin{subfigure}[b]{\gridColumnSpace}
            \centering
            \includegraphics[width=\textwidth]{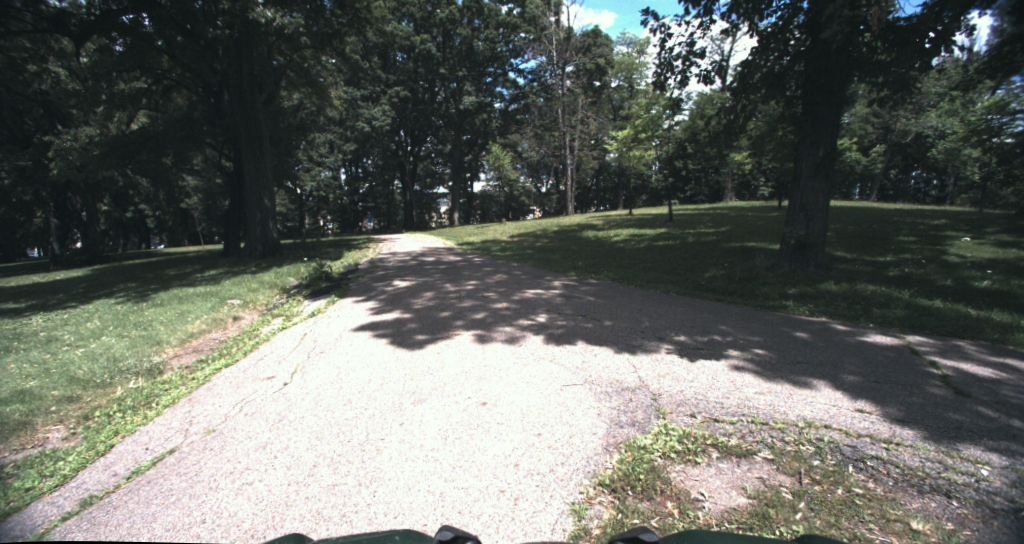}
        \end{subfigure} &
        \begin{subfigure}[b]{\gridColumnSpace}
            \centering
            \includegraphics[width=\textwidth]{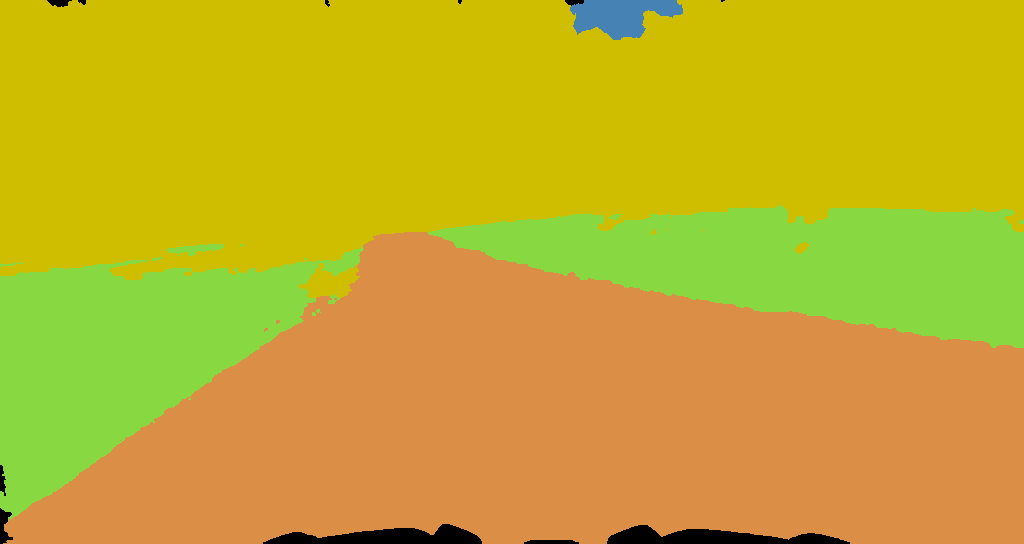}
        \end{subfigure} &
        \begin{subfigure}[b]{\gridColumnSpace}
            \centering
            \includegraphics[width=\textwidth]{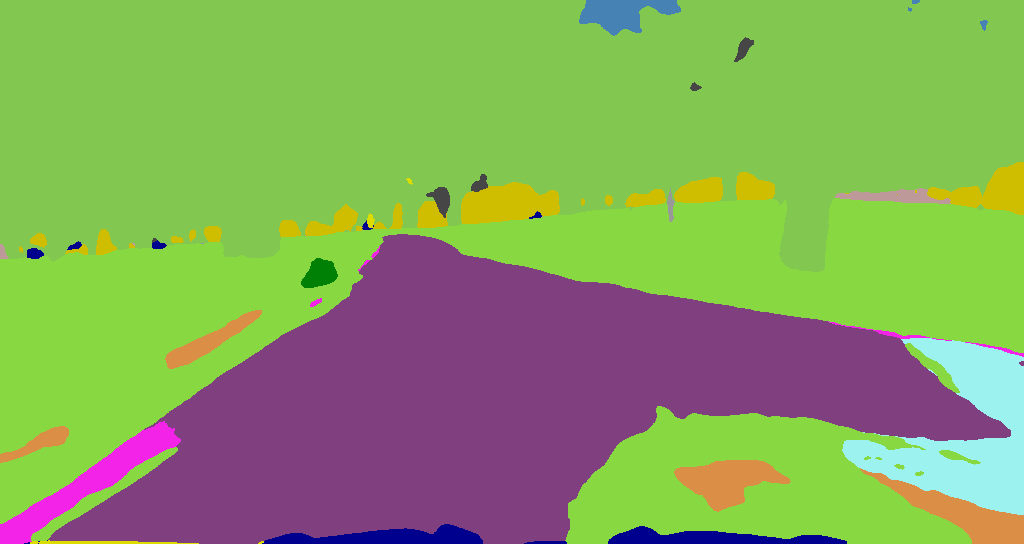}
        \end{subfigure} &
        \begin{subfigure}[b]{\gridColumnSpace}
            \centering
            \includegraphics[width=\textwidth]{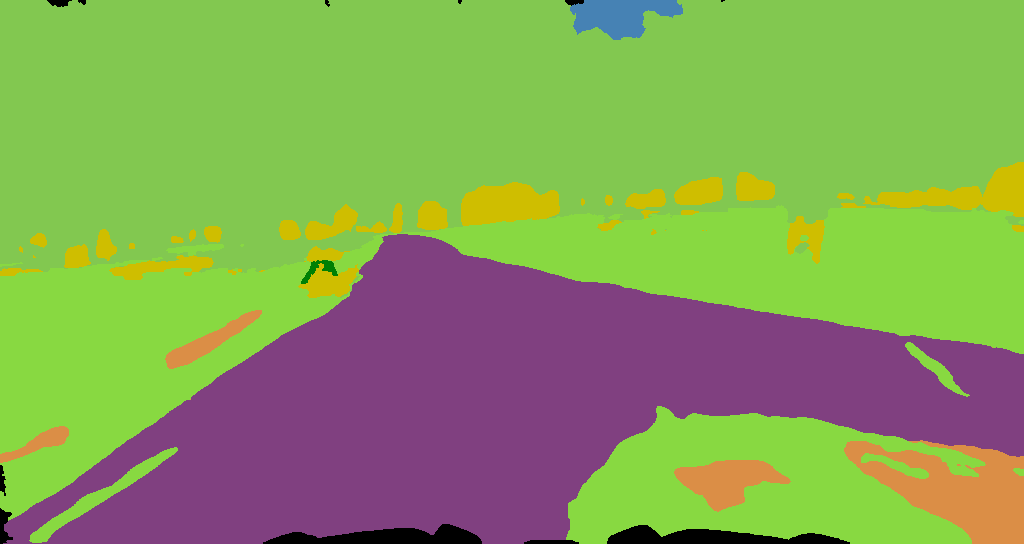}
        \end{subfigure}
        % \\[-1pt]
    \end{tabular}

    \caption{Pseudo-label creation on three datasets using the GOOSE source taxonomy which includes classes such as \texttt{sidewalk} (\textcolor[rgb]{0.9569, 0.1373, 0.9098}{\rule{1.5ex}{1.5ex}}), \texttt{cobblestone} (\textcolor[rgb]{0.6078, 0.949, 0.9373}{\rule{1.5ex}{1.5ex}}), \texttt{non-drivable vegetation} (\textcolor[rgb]{0.8118, 0.7451, 0.0}{\rule{1.5ex}{1.5ex}}), \texttt{low grass} (\textcolor[rgb]{0.5333, 0.851, 0.2549}{\rule{1.5ex}{1.5ex}}), \texttt{high grass} (\textcolor[rgb]{0.0, 0.502, 0.0157}{\rule{1.5ex}{1.5ex}}), \texttt{asphalt} (\textcolor[rgb]{0.502, 0.251, 0.502}{\rule{1.5ex}{1.5ex}}), and \texttt{rough drivable surface} (\textcolor[rgb]{0.8588, 0.5569, 0.2745}{\rule{1.5ex}{1.5ex}}).}
    \label{fig:quality_pics}
\end{figure*}

We evaluate our method using 7 datasets \cite{bib:yamaha, bib:rellis-3d, bib:cityscapes, bib:freiburg-forest, bib:tas500, bib:lanes} with GOOSE \cite{bib:goose-dataset} as the source dataset.
GOOSE consists of 8800 train and validation images.
The additional data consists of 9900 train images.
Off-road datasets differ from urban driving datasets in that they focus on the class of terrain.
This does not, however, exclude urban scenes, which is why Cityscapes is also included in training.
For off-road datasets there is no WildDash equivalent that can be used for evaluating robustness.
Hence, we evaluate only on the GOOSE validation split.

Table \ref{tab:goose-results} shows the accuracy improvement of the student model trained on the composite dataset measured by mean IoU.
Compared to the Cityscapes pre-trained model, the GOOSE model's improvements are less significant.
We largely attribute this to the low amount of \textit{diverse} data.
For example, the biggest contributor in terms of numbers is Rellis-3D with 6000 labeled frames.
Unfortunately, Rellis-3D only depicts 5 different scenes with each individual frame being recorded less than a meter apart.
Besides quality, off-road datasets also generally lack quantity compared to urban ones.
The unified off-road dataset is roughly twice the size of GOOSE, whereas the unified urban dataset is 36 times the size of Cityscapes.

% \subsection{Confirmation Bias}
% \label{chapter:confirmation-bias}
% plot some iteration 0, 1, 2 images to see how the inference changes over time

\subsection{Qualitative Results}
In addition to the quantitative analysis, we provide qualitative examples of pseudo-label generation in \cref{fig:quality_pics}.

\section{Conclusion and Future Work}
In this work, we proposed a novel approach for knowledge distillation that simultaneously addresses the challenge of label space unification in semantic segmentation. 
By leveraging dataset priors, we demonstrated how a teacher model trained on a source dataset can generate improved pseudo-labels for extra datasets, effectively reconciling inconsistent taxonomies. 
Our results show that student models trained on these refined pseudo-labels consistently outperform their teacher models across both urban and off-road domains.

Our experiments highlight key insights: (1) larger models benefit more from increased dataset diversity, (2) incorporating partially labeled but domain-shifted data yields better generalization than relying solely on pseudo-labeled data from the same dataset, and (3) dataset priors play a crucial role in improving pseudo-label accuracy. 
However, further iterations of self-training did not provide additional improvements.

Despite these promising results, several challenges remain open for future work. 
First, the impact of specific datasets on performance needs further investigation, particularly in cases where a large domain gap exists. 
A better understanding of when additional data becomes beneficial versus detrimental could lead to more efficient dataset selection strategies. 
Second, while iterative training and label refinement have shown promise in prior studies, our experiments yielded no improvements. This suggests that success may depend on factors like model architecture or implementation strategy.
Finding the optimal approach remains an open challenge and a promising avenue for future research.
Third, we observed that models trained on diverse datasets performed well on robustness benchmarks such as WildDash but did not always improve accuracy on the original source dataset (namely Mapillary-Vistas). 
Finding out which factors contribute to improving on the source dataset remains an open question.

%%%%%%%%%%%%%%%%%%%%%%%%%%%%%%%%%%%%%%%%%%%%%%%%%%%%%%%%%%%%%%%%%%%%%%%%%%%%%%%%

%%%%%%%%%%%%%%%%%%%%%%%%%%%%%%%%%%%%%%%%%%%%%%%%%%%%%%%%%%%%%%%%%%%%%%%%%%%%%%%%

\balance

\bibliographystyle{IEEEtran}
\bibliography{additional_abrv,IEEEabrv,et_al,backhaus-wahrnehmung}

\begin{thebibliography}{10}
\providecommand{\url}[1]{#1}
\csname url@rmstyle\endcsname
\providecommand{\newblock}{\relax}
\providecommand{\bibinfo}[2]{#2}
\providecommand\BIBentrySTDinterwordspacing{\spaceskip=0pt\relax}
\providecommand\BIBentryALTinterwordstretchfactor{4}
\providecommand\BIBentryALTinterwordspacing{\spaceskip=\fontdimen2\font plus
\BIBentryALTinterwordstretchfactor\fontdimen3\font minus \fontdimen4\font\relax}
\providecommand\BIBforeignlanguage[2]{{%
\expandafter\ifx\csname l@#1\endcsname\relax
\typeout{** WARNING: IEEEtran.bst: No hyphenation pattern has been}%
\typeout{** loaded for the language `#1'. Using the pattern for}%
\typeout{** the default language instead.}%
\else
\language=\csname l@#1\endcsname
\fi
#2}}

\bibitem{bib:apolloscape}
P.~Wang, \emph{et~al.}, ``{The ApolloScape Open Dataset for Autonomous Driving and its Application},'' \emph{IEEE transactions on pattern analysis and machine intelligence}, 2019.

\bibitem{bib:attention-is-all-you-need}
A.~Vaswani, \emph{et~al.}, ``{Attention Is All You Need},'' in \emph{{Advances in Neural Information Processing Systems (NIPS)}}, I.~Guyon, \emph{et~al.}, Eds., vol.~30.\hskip 1em plus 0.5em minus 0.4em\relax Curran Associates, Inc., 2017.

\bibitem{bib:a2d2}
J.~Geyer, \emph{et~al.}, ``{A2D2: Audi Autonomous Driving Dataset},'' 2020.

\bibitem{bib:bdd100k}
F.~Yu, \emph{et~al.}, ``{BDD100K: A Diverse Driving Dataset for Heterogeneous Multitask Learning},'' 2020.

\bibitem{bib:mapillary-vistas}
G.~Neuhold, \emph{et~al.}, ``{The Mapillary Vistas Dataset for Semantic Understanding of Street Scenes},'' in \emph{{Proc. IEEE Int. Conf. Comput. Vision (ICCV)}}, 2017, pp. 5000--5009.

\bibitem{bib:freiburg-forest}
A.~Valada, \emph{et~al.}, ``{Deep Multispectral Semantic Scene Understanding of Forested Environments using Multimodal Fusion},'' in \emph{{Int. Symp. Experimental Robotics (ISER)}}, 2016.

\bibitem{bib:yamaha}
D.~Maturana, \emph{et~al.}, ``{Real-time Semantic Mapping for Autonomous Off-Road Navigation},'' in \emph{Field and Service Robotics}.\hskip 1em plus 0.5em minus 0.4em\relax Springer, 2018, pp. 335--350.

\bibitem{bib:goose-dataset}
P.~Mortimer, \emph{et~al.}, ``{The GOOSE Dataset for Perception in Unstructured Environments},'' in \emph{{Proc. IEEE Int. Conf. Robotics and Automation (ICRA)}}, 2024.

\bibitem{bib:naive-student}
L.~Chen, \emph{et~al.}, ``Leveraging semi-supervised learning in video sequences for urban scene segmentation,'' \emph{CoRR}, vol. abs/2005.10266, 2020.

\bibitem{bib:billion-scale}
I.~Z. Yalniz, \emph{et~al.}, ``{Billion-scale semi-supervised learning for image classification},'' \emph{arXiv preprint arXiv:1905.00546}, 2019.

\bibitem{bib:noisy-student}
Q.~Xie, \emph{et~al.}, ``{Self-training with Noisy Student improves ImageNet classification},'' \emph{arXiv preprint arXiv:1911.04252}, 2019.

\bibitem{bib:mean-teacher}
A.~Tarvainen and H.~Valpola, ``{Mean teachers are better role models: Weight-averaged consistency targets improve semi-supervised deep learning results},'' in \emph{{Advances in Neural Information Processing Systems (NIPS)}}, vol.~30.\hskip 1em plus 0.5em minus 0.4em\relax Curran Associates, Inc., 2017.

\bibitem{bib:pp-liteseg}
J.~Peng, \emph{et~al.}, ``{PP-LiteSeg: A Superior Real-Time Semantic Segmentation Model},'' \emph{arXiv preprint arXiv:2204.02681}, 2022.

\bibitem{bib:meta-pseudo-labels}
H.~Pham, \emph{et~al.}, ``{Meta Pseudo Labels},'' in \emph{{Proc. IEEE Conf. Comput. Vision and Pattern Recognition (CVPR)}}, 2021, pp. 11\,552--11\,563.

\bibitem{bib:data-distillation}
I.~Radosavovic, \emph{et~al.}, ``{Data Distillation: Towards Omni-Supervised Learning},'' in \emph{{Proc. IEEE Conf. Comput. Vision and Pattern Recognition (CVPR)}}, 2018, pp. 4119--4128.

\bibitem{bib:rethinking-pretraining}
B.~Zoph, \emph{et~al.}, ``{Rethinking Pre-training and Self-training},'' in \emph{{Advances in Neural Information Processing Systems (NIPS)}}, H.~Larochelle, \emph{et~al.}, Eds., vol.~33.\hskip 1em plus 0.5em minus 0.4em\relax Curran Associates, Inc., 2020, pp. 3833--3845.

\bibitem{bib:depth-anything}
L.~Yang, \emph{et~al.}, ``{Depth Anything: Unleashing the Power of Large-Scale Unlabeled Data},'' in \emph{{Proc. IEEE Conf. Comput. Vision and Pattern Recognition (CVPR)}}, 2024, pp. 10\,371--10\,381.

\bibitem{bib:weakly-and-semi-supervised-learning}
G.~Papandreou, \emph{et~al.}, ``{Weakly- and Semi-Supervised Learning of a Deep Convolutional Network for Semantic Image Segmentation},'' in \emph{{Proc. IEEE Int. Conf. Comput. Vision (ICCV)}}, 2015, pp. 1742--1750.

\bibitem{bib:boosting-ssss}
P.~Meletis and G.~Dubbelman, ``{On Boosting Semantic Street Scene Segmentation with Weak Supervision},'' in \emph{{Proc. IEEE Intelligent Vehicles Symp. (IV)}}, 2019, pp. 1334--1339.

\bibitem{bib:bbam}
J.~Lee, \emph{et~al.}, ``{BBAM: Bounding Box Attribution Map for Weakly Supervised Semantic and Instance Segmentation},'' in \emph{{Proc. IEEE Conf. Comput. Vision and Pattern Recognition (CVPR)}}, 2021, pp. 2643--2652.

\bibitem{bib:urban-scene-coarse-annotation}
A.~Das, \emph{et~al.}, ``{Urban Scene Semantic Segmentation With Low-Cost Coarse Annotation},'' in \emph{{Proc. IEEE Winter Conf. Applicat. Comput. Vision (WACV)}}, 2023, pp. 5978--5987.

\bibitem{bib:image-level-weak-supervision}
J.~Ahn and S.~Kwak, ``{Learning Pixel-Level Semantic Affinity With Image-Level Supervision for Weakly Supervised Semantic Segmentation},'' in \emph{{Proc. IEEE Conf. Comput. Vision and Pattern Recognition (CVPR)}}, 2018.

\bibitem{bib:automated-label-unification}
R.~Ma, \emph{et~al.}, ``{Automated Label Unification for Multi-Dataset Semantic Segmentation with GNNs},'' in \emph{{Advances in Neural Information Processing Systems (NIPS)}}, 2024.

\bibitem{bib:multi-head-semseg}
S.~Masaki, \emph{et~al.}, ``{Multi-Domain Semantic-Segmentation using Multi-Head Model},'' in \emph{{Proc. IEEE Intelligent Transportation Syst. Conf. (ITSC)}}, 2021, pp. 2802--2807.

\bibitem{bib:cross-dataset-learning}
L.~Wang, \emph{et~al.}, ``{Cross-Dataset Collaborative Learning for Semantic Segmentation in Autonomous Driving},'' in \emph{{Proc. AAAI Conf. on Artificial Intelligence}}, vol.~36, no.~3, 2022, pp. 3180--3188.

\bibitem{bib:universal-ssss}
T.~Kalluri, \emph{et~al.}, ``{Universal Semi-Supervised Semantic Segmentation},'' in \emph{{Proc. IEEE Int. Conf. Comput. Vision (ICCV)}}, 2019, pp. 5259--5270.

\bibitem{bib:heterogeneous-street-datasets}
P.~Meletis and G.~Dubbelman, ``{Training of Convolutional Networks on Multiple Heterogeneous Datasets for Street Scene Semantic Segmentation},'' in \emph{{Proc. IEEE Intelligent Vehicles Symp. (IV)}}.\hskip 1em plus 0.5em minus 0.4em\relax Institute of Electrical and Electronics Engineers, 2018, pp. 1045--1050.

\bibitem{bib:multi-task-domain-learning}
D.~Fourure, \emph{et~al.}, ``{Multi-task, multi-domain learning: Application to semantic segmentation and pose regression},'' \emph{Neurocomputing}, vol. 251, pp. 68--80, 2017.

\bibitem{bib:universal-image-concepts}
P.~Bevandić, \emph{et~al.}, ``{Weakly supervised Training of universal visual concepts for multi-domain semantic segmentation},'' \emph{International Journal of Computer Vision}, vol. 132, no.~7, pp. 2450--2472, 2024.

\bibitem{bib:partial-label-losses}
J.~Cid-Sueiro, ``{Proper losses for learning from partial labels},'' in \emph{{Advances in Neural Information Processing Systems (NIPS)}}, F.~Pereira, \emph{et~al.}, Eds., vol.~25.\hskip 1em plus 0.5em minus 0.4em\relax Curran Associates, Inc., 2012.

\bibitem{bib:multi-domain-semseg}
P.~Bevandić, \emph{et~al.}, ``{Multi-Domain Semantic Segmentation with Overlapping Labels},'' in \emph{{Proc. IEEE Winter Conf. Applicat. Comput. Vision (WACV)}}, 2022, pp. 2615--2624.

\bibitem{bib:mseg}
J.~Lambert, \emph{et~al.}, ``{MSeg: A Composite Dataset for Multi-Domain Semantic Segmentation.}'' \emph{IEEE Trans. Pattern Anal. Mach. Intell.}, vol.~45, no.~1, pp. 796--810, 2023.

\bibitem{bib:unifying-off-road}
A.~Medellin, \emph{et~al.}, ``{Applications of Unifying Off-Road Datasets Through Ontology},'' in \emph{{Proc. Ground Vehicle Systems Engineering and Technology Symp. (GVSETS)}}, no. 2024-01-4074.\hskip 1em plus 0.5em minus 0.4em\relax Warrendale, PA: SAE International, 2024, pp. 1--13.

\bibitem{bib:heterogeneous-datasets}
P.~Meletis and G.~Dubbelman, ``{Training Semantic Segmentation on Heterogeneous Datasets},'' \emph{arXiv preprint arXiv:2301.07634}, 2023.

\bibitem{bib:mask2former}
B.~Cheng, \emph{et~al.}, ``Masked-attention mask transformer for universal image segmentation,'' in \emph{{Proc. IEEE Conf. Comput. Vision and Pattern Recognition (CVPR)}}, June 2022, pp. 1290--1299.

\bibitem{bib:oneformer}
J.~Jain, \emph{et~al.}, ``{OneFormer: One Transformer to Rule Universal Image Segmentation},'' in \emph{{Proc. IEEE Conf. Comput. Vision and Pattern Recognition (CVPR)}}, 2023, pp. 2989--2998.

\bibitem{bib:wilddash}
O.~Zendel, \emph{et~al.}, ``{WildDash - Creating Hazard-Aware Benchmarks},'' in \emph{{Proc. European Conf. Comput. Vision (ECCV)}}, 2018.

\bibitem{bib:nuscenes}
H.~Caesar, \emph{et~al.}, ``{nuScenes: A multimodal dataset for autonomous driving},'' in \emph{{Proc. IEEE Conf. Comput. Vision and Pattern Recognition (CVPR)}}, 2020.

\bibitem{bib:waymo-dataset}
P.~Sun, \emph{et~al.}, ``{Scalability in Perception for Autonomous Driving: Waymo Open Dataset},'' in \emph{{Proc. IEEE Conf. Comput. Vision and Pattern Recognition (CVPR)}}, 2020.

\bibitem{bib:idd}
S.~Dokania, \emph{et~al.}, ``{IDD-3D: Indian Driving Dataset for 3D Unstructured Road Scenes},'' in \emph{{Proc. IEEE Winter Conf. Applicat. Comput. Vision (WACV)}}, 2023, pp. 4482--4491.

\bibitem{bib:coco}
T.-Y. Lin, \emph{et~al.}, ``{Microsoft COCO: Common Objects in Context},'' 2014.

\bibitem{bib:cityscapes}
M.~Cordts, \emph{et~al.}, ``{The Cityscapes Dataset for Semantic Urban Scene Understanding},'' in \emph{{Proc. IEEE Conf. Comput. Vision and Pattern Recognition (CVPR)}}, 2016.

\bibitem{bib:camvid}
G.~J. Brostow, \emph{et~al.}, ``Semantic object classes in video: A high-definition ground truth database,'' \emph{Pattern Recognition Letters}, 2008.

\bibitem{bib:lanes}
N.~Azqalani, ``{Semantic Segmentation Makassar(IDN) Road Dataset},'' \url{https://www.kaggle.com/datasets/nublanazqalani/semantic-segmentation-makassaridn-road-dataset}, 2022, accessed: Nov 13, 2024.

\bibitem{bib:rellis-3d}
P.~Jiang, \emph{et~al.}, ``{RELLIS-3D Dataset: Data, Benchmarks and Analysis},'' 2020.

\bibitem{bib:tas500}
K.~A. Metzger, \emph{et~al.}, ``{A Fine-Grained Dataset and its Efficient Semantic Segmentation for Unstructured Driving Scenarios},'' in \emph{{Int. Conf. Pattern Recognition (ICPR)}}, Milano, Italy, 2021.

\bibitem{bib:ade20k}
B.~Zhou, \emph{et~al.}, ``{Semantic Understanding of Scenes Through the ADE20K Dataset},'' \emph{Int. J. Comput. Vis.}, vol. 127, no.~3, pp. 302--321, 2019.

\end{thebibliography}

\end{document}